\documentclass{llncs}
\usepackage{graphicx}
\usepackage{amsmath}
\begin{document}

\title{Benchmarking KAZE and MCM for Multiclass Classification}
\titlerunning{Object Classification}  
%
\author{Siddharth Srivastava \and Prerana Mukherjee
\and Brejesh Lall }
\authorrunning{Siddharth Srivastava et. al.} 
%
\tocauthor{Siddharth Srivastava, Prerana Mukherjee, Brejesh Lall}
\institute{Indian Institute of Technology,Delhi,India\\
\email{\{eez127506, eez138300,brejesh\}@ee.iitd.ac.in }\\
}

\maketitle              

\begin{abstract}
In this paper, we propose a novel approach for feature generation by appropriately fusing KAZE and SIFT features. We then use this feature set along with Minimal Complexity Machine(MCM) for object classification. We show that KAZE and SIFT features are complementary. Experimental results indicate that an elementary integration of these techniques can outperform the state-of-the-art approaches.

\end{abstract}
\section{Introduction}
Majority of computer vision research e.g. in the area of image classification, object recognition, localization, detection, segmentation, retrieval etc.involves objects in one form or the other. Development of algorithms in these areas is driven by trade-off between robustness and scalability. In this paper we focus on the very challenging problem of object classification. In recent times, machine learning techniques have found favour among researchers addressing the image classification problem.  Researchers are revisiting the deep learning tools and taking  permutations of feature sets to encompass the various characteristics of the images. The diverse nature of objects make it difficult to devise a single solution to handle all object classification problems. The challenges in this area of research can be attributed to the following factors:

\begin{itemize}
\item Number of classes
\item Number of instances of each class
\item Total number of images in the dataset
\item Relative ratios of training and testing images 
\item Intra-class variance due to clutter, pose variations, occlusion, illumination changes etc. 
\item Ground truth annotations.
\end{itemize}

These factors have an impact not only on the efficiency of the technique but also upon its complexity. Having a large number of classes poses the challenge of capturing the intra-class as well as inter-class variation precisely. The size of training data as well as the variation in it decides has a direct impact on the discriminability(and hence complexity) of the features chosen. On one hand, large data is good for training the classifer however, on the other hand it leads to the requirement of complex features. Also training with a small dataset results in the problem that it does not capture the variational changes, whereas, a larger dataset makes annotation difficult \cite{pinto2008real}. 

\begin{figure}[h]
    \centering
    \includegraphics[scale=0.8]{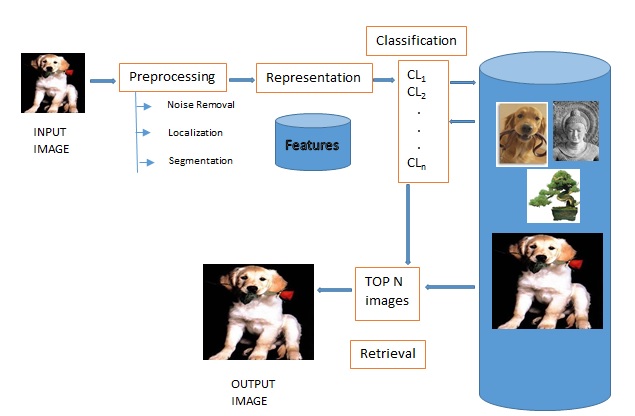}
    \caption{Object Classification Pipeline}
    \label{fig:img1}
\end{figure}

Various techniques attempt to address one or more of these factors for object classification. Broadly, the pipeline of such solutions may be described as in Figure \ref{fig:img1}.

Popular approaches for object classification can be categorized as follows:

\begin{enumerate}
\item Techniques which focus primarily on improving the input representation with the help of stronger features while using a simple classifier such as SVM. ScSPM proposed in  \cite{yang2009linear} chose sparse coding over vector quantization. Accordingly, it relaxes the cardinality constraints and introduces a regularization parameter to obtain a smaller number of non zero elements. This is then followed by max spatial pooling reducing the complexity of the classifier.  In \cite{wang2010locality}, authors use a locality adaptor which allows to choose appropriate basis vector corresponding to an input descriptor.

\item Techniques focusing on using classifiers by generating stronger training cases as compared to the approaches briefed in 1. In \cite{felzenszwalb2010object}, authors formulate a latent SVM which results in the problem being formulated as a convex training problem. They also propose a HoG like feature descriptor which is also used to generate hard training examples for building a stronger classifier. In \cite{girshick2014rich}, the authors introduce R-CNN, a variant of convolutional neural network to extract features from the region proposals which are then classified into respective object categories. 

\item Techniques which try to balance the trade-off between speed and accuracy by tuning both 1 and 2. In \cite{harzallah2009combining}, authors propose a two stage sliding window approach for object localization. The main idea is to combine the classification and detection phases by considering latent properties of objects and scenes. Another technique, Selective Search \cite{uijlings2013selective} reduces the relative time for localizing objects, hence allowing for stronger classification techniques.
\end{enumerate}

As is evident from the previous discussion, the choice of features and classifier play a crucial role in the quality of the object classification techniques, Despite this strong dependence on choice of features, SIFT \cite{lowe2004distinctive} and its variants \cite{van2010evaluating} \cite{ke2004pca} \cite{mortensen2005sift} have remained the de-facto choice for feature representation. SIFT is based on Gaussian scale space(GSS) which blurs the image uniformly, resulting in loss of distinctness in the object boundaries. Recently,a work proposes to use non linear scale space, which preserves the object boundaries by blurring the region around edges more than the edges themselves. KAZE \cite{alcantarilla2012kaze}, which is based on the non-linear scale space, hence is a promising choice for the features for object classification. KAZE features have strong responses around object boundaries, while SIFT features capture the details(at the boundary or otherwise) in an image. According to \cite{alexe2010object}, an object can be characterized by a well defined boundary, a distinctive appearance and a salient region. Therefore, a  feature set comprising of carefully chosen KAZE and SIFT keypoints is an appropriate choice for defining an object.

For the other key operations in object classification, Support Vector Machines(SVM) \cite{cortes1995support} have been the traditional choice. Recently, Minimal Complexity Machine (MCM) \cite{Jayadeva14} has shown to outperform SVM in terms accuracy, computational complexity as well as sparse representation of the features. The strongest argument in favor of MCM is its provably good generalization accuracy and requirement of far less number of support vectors as compared to SVMs. Fewer support vector mean faster classification of test points. Due to complexity and size of the object classification datasets, MCM makes a strong case for itself. A more detailed discussion about MCM as compared to SVM is given in Section 2.  

In light of the discussions above, we present a novel technique for generating a stronger feature set by careful combination of KAZE and SIFT keypoints(SIFT-KAZE).   We use these features with MCM to propose a light weight but stronger object classifier.
 The contributions of this paper can be summarized as follows:
\begin{enumerate}
\item This paper establishes that SIFT and KAZE are complementary features and a tuned combination of these is better suited for object classification tasks.  
\item This is the first work to demonstrate the effectiveness of MCM on images and datasets with large number of classes. 
Further, the proposed technique outperforms the traditional methods by a significant margin and can be easily integrated with the existing techniques. This can lead to the development of more efficient yet simpler techniques in this domain.
\end{enumerate}

Rest of the paper is organized as follows: In Section 2, we introduce the fundamental analysis of the non linear scale space and demonstrate its effectiveness in combination to the object boundary representation and go on to propose the object classification technique. Section 3, presents the experimental analysis. Section 4, elaborates the results which were obtained. Section 5, concludes the paper.

\section{Discriminant Keypoint based Classifier}

\subsection{Beyond SIFT}
The major difference between KAZE and SIFT is in the construction of the scale space. KAZE is based on non-linear scale space while SIFT is based upon Gaussian scale space(GSS). KAZE uses non-linear diffusion filtering. This diffusion process is formulated in equation (\ref{eq:1})
\begin{equation}\label{eq:1}
\frac{\partial L}{\partial t}=div\left \{ \left ( c\left ( x,y,t \right ) .\nabla\left ( L \right )\right ) \right \}
\end{equation}
where div and $\nabla$ are divergence and gradient operators, c is the conductivity function and t is scale parameter.
The conductivity function c, is represented as a gradient(Equation (\ref{eq:3}), helping in the reduction of diffusion at edges resulting in more smoothening of regions than edges. This property of the conductivity function makes it more suitable for boundary representation. 

SIFT constructs GSS which blurs both the object region and boundary. This helps in characterization of object using high detail interest points(not necessarily boundary points). KAZE uses the general diffusion equation as shown in equation (\ref{eq:3}) to construct the scale space. There are various conductivity functions defined in \cite{perona1990scale}, which can be used to promote high contrast, wider regions or  smoothening on both sides of the edges.
 
\begin{equation}\label{eq:3}
c\left ( x,y,t \right )=g\left ( \left | \nabla L_{\sigma }\left ( x,y,t \right ) \right | \right )
\end{equation}
In SIFT,the base image for each octave is generated by downsampling the image from previous octave whereas in KAZE, the construction of each octave is based on the original image .
\begin{figure}[h]
    \centering
    \includegraphics[scale=0.8]{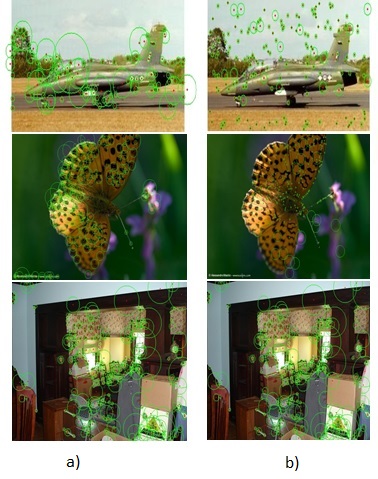}
    \caption{a) Shows the KAZE keypoints which are densely distributed along he object boundaries and b) Shows the SIFT keypoints around the regions.}
    \label{fig:img2}
\end{figure}

\begin{figure}[h]
    \centering
    \includegraphics[scale=0.8]{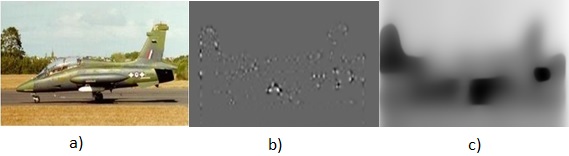}
    \caption{a) Original Image b) KAZE detector response localised around the object (aeroplane) c) KAZE scalespace }
    \label{fig:img3}
\end{figure}

\begin{figure}[h]
    \centering
    \includegraphics[scale=0.5]{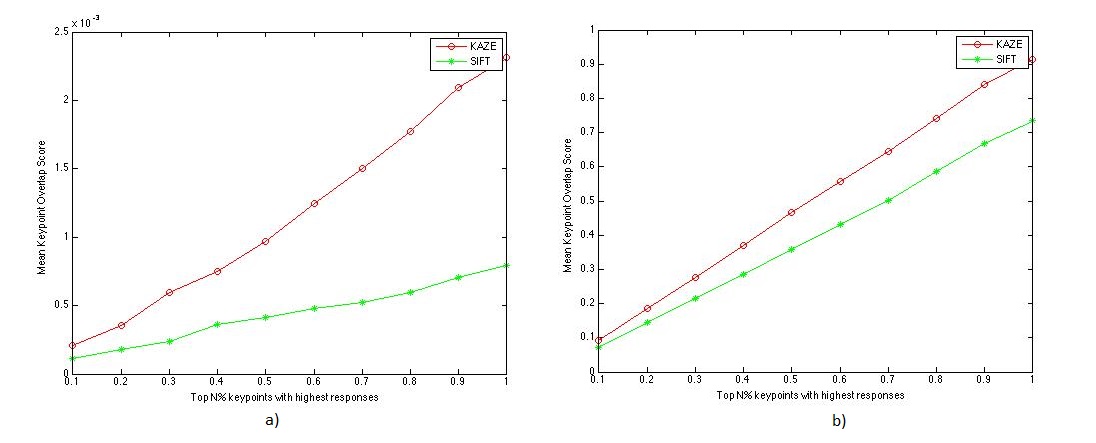}
    \caption{(a) Mean Average Keypoint Overlap Score vs Top N\% keypoints with highest responses(For all keypoints within the bounding box) (b) Mean Average Keypoint Score vs Top N\% keypoints with highest responses (For all keypoints within  $\beta$ = 0.1)}
    \label{fig:img4}
\end{figure}

We now define two measures Keypoint Overlap Score (KOS) and Mean Keypoint Overlap Score (MKOS) to evaluate the effectiveness of SIFT and KAZE in providing discriminative keypoints.

The Keypoint Overlap Score(Equation \ref{eq:4}) of an image I is defined as the percentage of the number of keypoints within  the ground truth bounding boxes $BB_o$ for each object $o$ in the image. 
\begin{equation}\label{eq:4}
KOS = \frac{1}{K}\left [ \sum_{o=1}^{O} \sum_{k=1}^{K}\chi \left ( BB_{o} ,KP_{k}\right )\right ]
\end{equation}
\\
where
$O$ is the number of objects in the image,
$K$ is the total number of keypoints detected in the image,
$BB_{o}$ is the bounding box of object $o$,
$KP_{k}$ is the $k^{th}$ keypoint and 
$\chi \left ( BB_{o} ,KP_{k}\right )$ specifies if a keypoint $KP_{k}$ lies within the bounding box $BB_{o}$ (Equation \ref{eq:5}). 

\begin{equation}\label{eq:5}
\chi \left ( BB_{o} ,KP_{k}\right ) = 
\left\{
	\begin{array}{ll}
		1  & \quad \mbox{if } KP_{k} \enspace \mbox{within} \enspace  BB_{o} \\
		0 & \quad \mbox{otherwise} 
	\end{array}
\right.
\end{equation}

\begin{equation}\label{eq:6}
MKOS=\frac{1}{N}\sum_{i=1}^{N}KOS_{i}
\end{equation}
\\
where
$KOS_{i}$ is the Keypoint Overlap Score for Image $i$ and
$N$ is the total number of images \\

Since KOS is image specific, we define a generic goodness measure MKOS as the average over all the images considered for evaluation(Equation \ref{eq:6}).

The KOS and MKOS are calculated on Pascal VOC 2007 \cite{pascal-voc-2007} dataset.  To characterize the boundaries from the ground truth annotations, we also create a region around the ground truth box $BB_{o}$  by extending and reducing it with a factor of $\beta$ as shown in equations (\ref{eq:7}) and (\ref{eq:8}). The scores are then calculated for the region represented by $A_{region}$.

\begin{equation}\label{eq:7}
A_{extended}\left\lbrace BB_{o}\right\rbrace = A_{original}\left\lbrace BB_{o}\right\rbrace * (1+ \beta)
\end{equation}

\begin{equation}\label{eq:8}
A_{reduced}\left\lbrace BB_{o}\right\rbrace = A_{original}\left\lbrace BB_{o}\right\rbrace * (1-\beta)
\end{equation}

\begin{equation}\label{eq:9}
A_{region}\left\lbrace BB_{o}\right\rbrace = A_{extended}\left\lbrace BB_{o}\right\rbrace \cap A_{reduced}\left\lbrace BB_{o}\right\rbrace
\end{equation}

Figure \ref{fig:img4}(a) and (b) shows the MKOS score for this dataset. The results were produced by calculating KAZE and SIFT keypoints and responses for each image in the dataset and sorting them according to the responses of the keypoints. The MKOS was then calculated for top N\% of the keypoints.  As it is shown, KAZE consistently outperforms SIFT with the density of KAZE features being heavily concentrated around the object boundaries. Though representation of object boundary is an important factor for object classification, it is not the sole discriminating factor which follows from the experimental analysis in Section 3. Analytically non-linear scale space preserves edges and hence it is not surprising that most of the KAZE keypoints are concentrated at the boundary. SIFT on the other hand looks for sharp discontinuities at all scales and can hence capture keypoints inside a region. This phenomena can also be observed visually from Figure \ref{fig:img2}, SIFT gives a high number of keypoints in relatively less salient regions (like grass, clouds etc.), while KAZE features were dominant around the most salient region boundaries (i.e. the object boundaries). It can been observed in Figure \ref{fig:img3} that KAZE does not blurs out the object in the detector response as well as scale space. 
\subsection{SVM vs MCM}
Minimal Complexity Machine is based on bounding the Vapnik-Chervonenkis (VC) dimension. VC dimension is a measure to establish the effectiveness of a machine learning algorithm. Alternatively, consider a parametric model $M(\alpha)$ and a set of data points $\boldsymbol X$. Now if there exists a parameter $\alpha$ for model $M$, such that all possible  label assignments $\boldsymbol L$ to $\boldsymbol X$ are classified without errors. The model $M(\alpha)$ is then said to have shattered the data points $\boldsymbol X$. The largest number of data points that can be shattered by $M(\alpha)$ is defined as the VC dimension of this model. Thus, VC dimension gives a family of functions that separates the input set of points. More intuitively, VC dimension therefore sets an upper bound on the test error rate\cite{vapnik1998statistical}.  The performance of the model is evaluated by risk associated with it, which is given as 

\begin{equation}\label{eq:risk}
Risk \leq Empirical Risk + f(h)
\end{equation}
where h is the VC dimension.

Here, the empirical risk is the classification error rate while $f$ is a monotonically increasing function.

Now as noted by Burges\cite{burges1998tutorial}, SVMs may have a very high VC dimension. Consequently, it would have a high risk associated it as compared to a model with the same classification error rate. In contrast, MCM guarantees good generalization  accuracy by obtaining a better bound (lower and upper) over the VC dimension while also achieving excellent training error rates. In addition as noted in \cite{Jayadeva14}, the number of support vectors obtained by MCM are comparatively less than that of SVM. This makes MCM suitable for complex classification tasks while also providing opportunity to reduce the overall overhead during classification. Since MCM solves a linear programming problem, it provides a significant performance gain over the quadratic programming problem solved by SVM.

\section{Experiments and Results}
In the experiments we evaluated SIFT, KAZE and the proposed SIFT-KAZE against each other using SVM and MCM as classifiers. We have provided an exhaustive evaluation over Caltech-256 dataset\cite{griffin2007caltech}.

We represent the features as bag of visual words which is then provided to SVM and MCM for further classification. For multiclass classification, We have used the one-vs-one approach and libSVM\cite{CC01a} implementation for SVM classification. As the patterns represented by SIFT features are linearly separable \cite{yang2009linear}, a linear kernel is more suitable for classification. Therefore, for our experiments, we have used linear kernel instead of a non-linear kernel. We have found that the patterns represented by KAZE features are also linearly separable, since our experiments with a non linear(RBF) kernel were consistently performed worse than those with linear kernel.

The results are shown in Tables 1,2 and 3. MCM outperforms SVM in all the experiments. Here, it is important to reiterate that the contemporary works achieving state of the art performance using SVM used strong pre-processing techniques or were trained with specifically constructed hard negatives from the training examples whereas in this work, we have used the simplest representation of features and classifiers.

The comparatively lower classification accuracy of KAZE can be attributed to the fact that KAZE features have a high density along the boundary of the objects. This establishes that despite the fact that boundary is the strongest distinguishing property of an object, it is not the definitive criteria \cite{alexe2010object}. On the contrary, the weighted mixture of SIFT and KAZE (Table 3) outperform the other two approximately by 2-3\% for MCM and around 8\%-10\% for SVM. This strengthens the claim that SIFT and KAZE are complementary features and can strongly define an object within an image. This can be understood by observing the  fact that while KAZE effectively incorporates the boundary characteristics, SIFT prominently captures the region properties. 

Table 4, shows the performance of the state of the art technique on Caltech-256 dataset. As can be seen that only the CNN using ImageNet(pretrained) \cite{zeiler2014visualizing} outperforms our method. It is important to note that despite using the most basic technique, we were able to outperform many advanced and relatively complex techniques while also achieving comparable results to the state of the art. This is a key observation since the presented set of techniques are generic and hence numerous variants may be derived similar to contemporary techniques utilizing SVM and SIFT with myriad kind of tunings, preprocessing etc. 
\begin{table}[h] \label{tab:siftclassification}
\caption{Classification accuracy for MCM and SVM for SIFT features on Caltech-256 dataset.}
\begin{center}
\begin{tabular}{r@{\quad}rl}
\hline
Training Samples & MCM   & \quad SVM   \\
\hline\rule{0pt}{12pt}
15            & 52.79 & \quad 19.82 \\
30            & 55.08 & \quad 26.82 \\
45            & 56.45 & \quad 28.98 \\
60            & 57.20 & \quad 30.91 \\
\hline     
\end{tabular}
\end{center}
\end{table}

\begin{table}[h]\label{tab:kazeclassification}
\caption{Classification accuracy for MCM and SVM for KAZE features on Caltech-256 dataset.}
\begin{center}
\begin{tabular}{r@{\quad}rl}
\hline
Training Samples & MCM   & \quad SVM   \\
\hline\rule{0pt}{12pt}
15            & 51.83 & \quad 18.24 \\
30            & 52.00 & \quad 21.08 \\
45            & 52.70 & \quad 22.86 \\
60            & 52.90 & \quad 24.23 \\
\hline     
\end{tabular}
\end{center}
\end{table}

\begin{table}[h]\label{tab:siftkazeclassification}
\caption{Classification accuracy for MCM and SVM for Mixture of SIFT and KAZE features on Caltech-256 dataset.}
\begin{center}
\begin{tabular}{r@{\quad}rl}
\hline
Training Samples & MCM   & \quad SVM   \\
\hline\rule{0pt}{12pt}
15            & 56.93 & \quad 26.86 \\
30            & 57.13 & \quad 34.92 \\
45            & 58.68 & \quad 38.95 \\
60            & 59.66 & \quad 42.60 \\
\hline     
\end{tabular}
\end{center}
\end{table}

\begin{table}[h]
\caption{State of the art classification accuracy on Caltech-256}
\begin{center}

\begin{tabular}{lllll}

\hline
 Technique \quad & 15 \quad & 30 \quad & 45 \quad & 60 \quad \\
 \hline
 ScSPM[2009]\cite{yang2009linear} & 27.73 \quad  & 34.02 \quad  & 37.46 \quad &  40.14 \quad \\
 LLC[2010] \cite{wang2010locality} & 34.36 \quad & 41.19 \quad & 45.31 \quad & 47.68 \\
 Multipath Sparse Coding[2012] \cite{bo2013multipath}  & 40.5 \quad & 48.0 \quad & 51.9 \quad & 55.20 \\
SIFT+Fisher Vector[2013]\cite{sanchez2013image} & 38.5 \quad & 47.4 \quad & 52.1 \quad & 54.8 \\
 SIFT+LCS+Fisher Vector[2013]\cite{sanchez2013image} & 41.0 \quad & 49.4 \quad & 54.3 \quad & 57.3 \\
CNN using ImageNet pretrained[2014]\cite{zeiler2014visualizing} & 65.7 \quad & 70.6 \quad & 72.7 \quad & 74.2 \\
\hline
\end{tabular}

\end{center}
\end{table}

\clearpage
\section{Conclusion and Future Scope}

We have established that SIFT and KAZE features represent complementary information of an object and a fusion of these techniques along with MCM outperforms the state of the art, while achieving remarkable improvement over SVM classification. We also evaluated the effectiveness of MCM for image datasets.The set of techniques used in this paper are simple yet powerful, we trust that they have the potential to significantly improve the more sophisticated(complex) state of the art techniques.

\bibliographystyle{abbrv}
\bibliography{sample}
\end{document}